\documentclass[sigconf]{acmart}
\AtBeginDocument{%
  \providecommand\BibTeX{{%
    \normalfont B\kern-0.5em{\scshape i\kern-0.25em b}\kern-0.8em\TeX}}}

\copyrightyear{2022}
\acmYear{2022}
\setcopyright{acmcopyright}
\acmConference[WSDM '22]{Proceedings of the Fifteenth ACM International Conference on Web Search and Data Mining}{February 21--25, 2022}{Tempe, AZ, USA}
\acmBooktitle{Proceedings of the Fifteenth ACM International Conference on Web Search and Data Mining (WSDM '22), February 21--25, 2022, Tempe, AZ, USA}
\acmPrice{15.00}
\acmDOI{10.1145/3488560.3498387}
\acmISBN{978-1-4503-9132-0/22/02}

\usepackage{algorithm,algorithmicx,algpseudocode}
\usepackage{multirow}
\usepackage{subfigure}

\acmSubmissionID{150}

\begin{document}

\title{Translating Human Mobility Forecasting through Natural Language Generation}
\author{Hao Xue}
\email{hao.xue@rmit.edu.au}
\affiliation{%
  \institution{School of Computing Technologies, RMIT University}
  \city{Melbourne}
  \state{Victoria}
  \country{Australia}
}

\author{Flora D. Salim}
\email{flora.salim@rmit.edu.au}
\affiliation{%
  \institution{School of Computing Technologies, RMIT University}
  \city{Melbourne}
  \state{Victoria}
  \country{Australia}
}

\author{Yongli Ren}
\email{yongli.ren@rmit.edu.au}
\affiliation{%
  \institution{School of Computing Technologies, RMIT University}
  \city{Melbourne}
  \state{Victoria}
  \country{Australia}
}

\author{Charles L. A. Clarke}
\email{charles.clarke@uwaterloo.ca}
\affiliation{%
  \institution{University of Waterloo}
  \city{Waterloo}
  \state{Ontario}
  \country{Canada}
}

\newcommand{\name}{SHIFT}
\newcommand{\pname}{S2S}
\newcommand{\mb}{\textit{Mob}}
\newcommand{\nl}{\textit{NL}}
\newcommand{\eg}{\textit{e.g.}}
\newcommand{\ie}{\textit{i.e.}}
\newcommand{\et}{\textit{et al.}}
\begin{abstract}

Existing human mobility forecasting models follow the standard design of the time-series prediction model which takes a series of numerical values as input to generate a numerical value as a prediction. Although treating this as a regression problem seems straightforward, incorporating various contextual information such as the semantic category information of each Place-of-Interest (POI) is a necessary step, and often the bottleneck, in designing an effective mobility prediction model. As opposed to the typical approach, we treat forecasting as a translation problem and propose a novel forecasting through a language generation pipeline. The paper aims to address the human mobility forecasting problem as a language translation task in a sequence-to-sequence manner. A mobility-to-language template is first introduced to describe the numerical mobility data as natural language sentences. The core intuition of the human mobility forecasting translation task is to convert the input mobility description sentences into a future mobility description from which the prediction target can be obtained. Under this pipeline, a two-branch network, SHIFT (Tran\textbf{s}lating \textbf{H}uman Mob\textbf{i}lity \textbf{F}orecas\textbf{t}ing), is designed. Specifically, it consists of one main branch for language generation and one auxiliary branch to directly learn mobility patterns. During the training, we develop a momentum mode for better connecting and training the two branches. Extensive experiments on three real-world datasets demonstrate that the proposed SHIFT is effective and presents a new revolutionary approach to forecasting human mobility. 

\end{abstract}




\maketitle

\section{Introduction}

Human mobility forecasting such as the next location prediction task~\cite{feng2018deepmove,miao2020predicting} and the customer flow prediction task~\cite{ma2020forecasting} are  essential ingredients in many domains including human mobility understanding and smart city applications. During the pandemic, support for contact tracing and crowd management has become of critical importance.
In the literature, human mobility prediction is invariably addressed through a time-series forecasting framework.
In this forecasting framework, the model takes numerical mobility data (\eg, number of visits of each POI) as input and yields a predicted value for some future time period.

The last few years have seen major advances in deep learning techniques for mobility prediction models. In order to better capture human mobility patterns and predict future mobility, these models consider different types of contextual information beyond historical mobility records. 
For example, the semantic category associated with points of interest (POIS) are incorporated into human mobility prediction models~\cite{chen2020context,guo2020attentional,feng2020predicting} and external information, such as local weather conditions, day of the week, and time, are incorporated into traffic flow prediction models~\cite{zhang2019flow,termcast,miao2021predicting}.

\begin{figure}[!htbp]
    \centering
    \includegraphics[width=.3\textwidth]{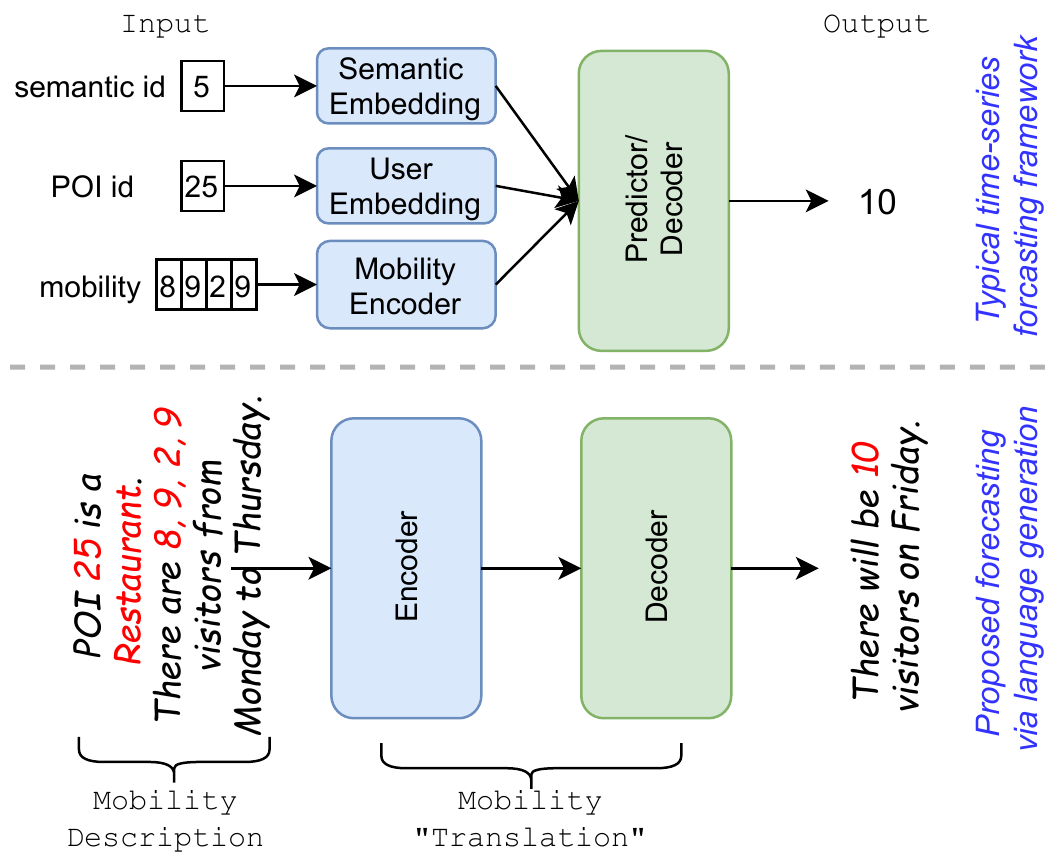}
    \caption{Conceptual illustrations of the typical time-series forecasting framework for human mobility prediction (the upper half) and the proposed forecasting through language generation (the lower half).}
    \label{fig:intro}
\end{figure}

This prediction workflow is simplified and summarized in the upper half of Figure~\ref{fig:intro}.
Multiple data sources providing diverse information (\eg, POI id, semantic information, and the historical mobility) are first passed through several encoders or embedding layers to extract feature vectors.
After the encoding process, the extracted/embedded contextual features are concatenated as the input of a predictor/decoder to yield the prediction (\eg, \textit{10} in the example given in the figure).
There are two major limitations of this framework:
(a) When there are multiple contexts, the concatenation operation may not be the optimal way of merging different data sources. It may be difficult to learn or capture the latent correlations of multiple contexts when multiple features are appended together.
(b) Considering the inherent characteristics of different contexts, several different feature encoders or embedding layers are necessary for a prediction model to learn the influence of these contexts.
These additions may dramatically increase the complexity of the prediction model and makes the model harder to train.

Inspired by the development of natural language processing models, we notice that a neural machine translation structure could be a suitable solution to address the above limitations.
Assuming that all types of contextual information and data sources can be described in a natural language sentence, the prediction model then needs only to take the sentence as input without utilizing different encoders or worrying how to combine different contexts.
In this paper, we seek to answer the research question: \textit{Can we predict human mobility in a natural language translation manner while maintaining a higher mobility forecasting performance?}

Unlike existing methods for human mobility prediction, we create an unconventional pipeline for mobility translation, essentially translating from  historical mobility to future mobility.
As illustrated in the lower part of Figure~\ref{fig:intro}, our proposed method for forecasting via a language generation pipeline is a sequence-to-sequence structure. Through a mobility description, both mobility data and other supporting contextual information are transformed into natural language sentences.
Then, in a mobility translation step, these descriptive sentences are taken as input and a natural language sentence indicating the prediction is generated as output.

Specifically, in this paper we propose a novel two-branch architecture, \name\ (Tran\textbf{s}lating \textbf{H}uman Mob\textbf{i}lity \textbf{F}orecas\textbf{t}ing), for the core mobility translation part under the above human mobility forecasting through language generation pipeline.
The architecture consists of a main natural language branch (\nl) and an auxiliary mobility branch (\mb).
The \nl\ branch is implemented as a sequence-to-sequence structure to ``translate" mobility descriptions, whereas the \mb\ branch focuses on learning mobility patterns.
The purpose of the auxiliary branch is to further improve the main branch's ability to generate mobility predictions.

In summary, our contributions are three-fold:
\begin{itemize}
    \item We explore and develop a novel mobility forecasting method through a language generation pipeline.
    To the best of our knowledge, this is the first work that addresses human mobility prediction task (time-series data forecasting) from the perspective of natural language translation and generation.
    \item We propose a two-branch network, \name, which has a main branch for language generation and an auxiliary branch for explicitly learning mobility patterns. To connect the two branches, a momentum averaging-based method is also introduced. 
    \item We conduct extensive experiments on three real-world datasets. The results demonstrate the superior performance of our \name\ and the effectiveness of each of its components.
\end{itemize}

\section{Related Work}
For human mobility prediction, existing methods can be categorized into two types: classical methods and deep learning-based methods. 
Under the classical category, based on ARIMA and Seasonal ARIMA, different methods have been designed~\cite{williams2003modeling,li2012prediction,lippi2013short} for forecasting crowd flow.
In addition, the Matrix Factorization is widely applied in next POI recommendation methods including~\cite{lian2014geomf,li2015rank,ren2017context}.

Deep learning-based methods mostly leverage Recurrent Neural Networks (RNNs) (as well as Long Short Term Memory (LSTM) networks~\cite{hochreiter1997long} and Gated Recurrent Units (GRU)~\cite{chung2014empirical}) for capturing mobility patterns in the observation history sequences.
ST-RNN proposed by Liu~\et~\cite{liu2016predicting} and DeepMove~\cite{feng2018deepmove} are two popular methods that apply attention mechanisms upon RNN to forecast human mobility.
Following this trend, various methods~\cite{chen2020context,guo2020attentional,yang2020location,luo2021stan} incorporating different context information have been proposed to predict human mobility.
Since the introduction of the self-attention-based Transformer architecture~\cite{vaswani2017attention}, it has been applied and achieved great success in many fields such as computer vision~\cite{vit,arnab2021vivit}, audio processing~\cite{xue2021exploring}, and natural language processing~\cite{bert,lewis2020bart}.
Recently, based on the effective Transformer, various methods have been designed and introduced for time-series data forecasting~\cite{li2019enhancing,zhou2021informer,wu2021autoformer} and specifically for human mobility prediction~\cite{halder2021transformer,termcast,wu2020hierarchically}.

Compared to the above-reviewed human mobility forecasting methods, we reshape the human mobility forecasting task and aim to address the prediction from the perspective of language translation.
The proposed method designed for translating human mobility forecasting in this work differs from these existing techniques.

\section{Method}

\subsection{Problem Formulation}

Assume that there is a set of POIs (place-of-interests) in a city: $\mathbf{U} = \{{u}_1, {u}_2, \cdots, {u}_{p}\}$.
For each POI $u$, $c_u$ stands for the semantic category information, such as a restaurant or a park.
The number of visits in a day $t$ of POI $u$ is represented as $x_t^u$.
The human mobility forecasting problem focused in this work is defined as follows.
Given the history record of visiting numbers $\mathbf{X}^u = [x_{t_1}^u, x_{t_2}^u, \cdots,x_{t_{\text{obs}}}^u]$, the goal is to predict the number of visits $\hat{x}_{t_{\text{obs} + 1}} ^u$ for the next day $t_{\text{obs} + 1}$.
The ground truth of visiting number is represented as $x_{t_{\text{obs} + 1}} ^u$ and ${\text{obs}}$ stands for the observation length of the given history visiting record.
For simplification, the superscript $u$ (indicating the POI id) is ignored in the rest of the paper.


\subsection{Mobility Description}\label{sec:des}
\begin{table*}[]
\centering
\caption{The template of mobility-to-language description. In the proposed mobility forecasting through language generation pipeline, the input part can be seen as the source sentences and the output part is the destination sentence.}
\addtolength{\tabcolsep}{-0.85ex}
\begin{tabular}{|l|l|p{2.35in}|p{2.5in}|} \hline
 & Description       & Template & Example                       \\ \hline\hline
\multirow{4}{*}{Input} & POI Semantic       & Place-of-Interest (POI) \{$u$\} is a/an \{$c_u$\}.            & Place-of-Interest (POI) 81 is a Optical Goods Store.              \\ \cline{2-4}
                       & Observation Time   & From \{$t_1$\} to \{$t_{\text{obs}}$\},                                    & From August 26, 2020, Wednesday to August 28, 2020, Friday,          \\\cline{2-4}
                       & Mobility Data      & there were \{$[x_{t_1}, x_{t_2}, \cdots,x_{t_{\text{obs}}}]$\} people visiting POI \{$u$\} on each day. & there were 42, 32, 29 people visiting POI 81 on each day. \\\cline{2-4}
 & Prediction Target Time & On \{$t_{\text{obs}+1}$\}, & On August 29, 2020, Saturday, \\ \hline
Output                 & Prediction Results & there will be \{${x}_{t_{\text{obs} + 1}}$ \} people visiting POI \{$u$\}.          & there will be 21 people visiting POI 81.  \\ \hline                         
\end{tabular}
\label{tab:mob_language}
\end{table*}

In the proposed forecasting via language generation pipeline, an important step to be addressed is how to describe the mobility data (always available in the numerical format) in natural language.
This mobility-to-language transformation provides the source sentences and the destination sentences for the ``mobility translator".
Given that this work explores how to leverage language models for mobility prediction for the first time, to the best of our knowledge, there is no prior work available for mobility-to-language transformation.
Therefore, we first develop a simple yet effective template-based method for mobility description.

Table~\ref{tab:mob_language} demonstrates the mechanism of the proposed mobility description method. 
Generally, there are two parts included: input description generation and output description generation. 
For the input description, it produces \textit{prompts} that serve as the input natural language sentences of the encoder (blue box in the lower half of Figure~\ref{fig:intro}).
As given in the table, the prompt consists of four elements:
\begin{itemize}
    \item POI Semantic: to give the POI id and describe the semantic category information of the POI;
    \item Observation Time: to indicate the timestamps of the observation period;
    \item Mobility Data: to transform the numerical mobility data into natural language, which is the essential part of the prompt;
    \item Prediction Target: to provide a cue of the prediction target timestamp $t_{\text{obs}+1}$.
\end{itemize}
By linking all four elements together (the first four rows in Table~\ref{tab:mob_language}), the entire prompt is then generated.

Similarly, the output description part (used as the ground truth for training and evaluation) handles the targeting sentences which are the expected output of the decoder (green box in the lower half of Figure~\ref{fig:intro}).
It has only one sentence and focuses on the prediction goal $x_{t_{\text{obs} + 1}}$.
One example of the output description is given in the last row of Table~\ref{tab:mob_language}.

Depending on the available data or the application, other sentences for describing extra information for mobility prediction such as holiday information (\eg, {\fontfamily{qcr}\selectfont Tuesday is Boxing Day.}) and weather conditions (\eg, {\fontfamily{qcr}\selectfont There were showers on Thursday.}) could also be easily appended in the prompt.
For the conventional time-series forecasting frameworks, in order to take various types of extra information into consideration, it is necessary to explicitly design and introduce extra modules or layers such as the external component in~\cite{zhang2017deep} and the gating mechanism to fuse external information in~\cite{zhang2019flow}.
On the contrary, the proposed language generation-based mobility prediction method only needs to update the prompts instead of adding extra layers or tweaking the model architecture.
This reflects the flexibility of the proposed forecasting via language generation pipeline.

\begin{figure*}
    \centering
    \subfigure[Basic mode]{\label{fig:basic}
    \includegraphics[width=.32\textwidth]{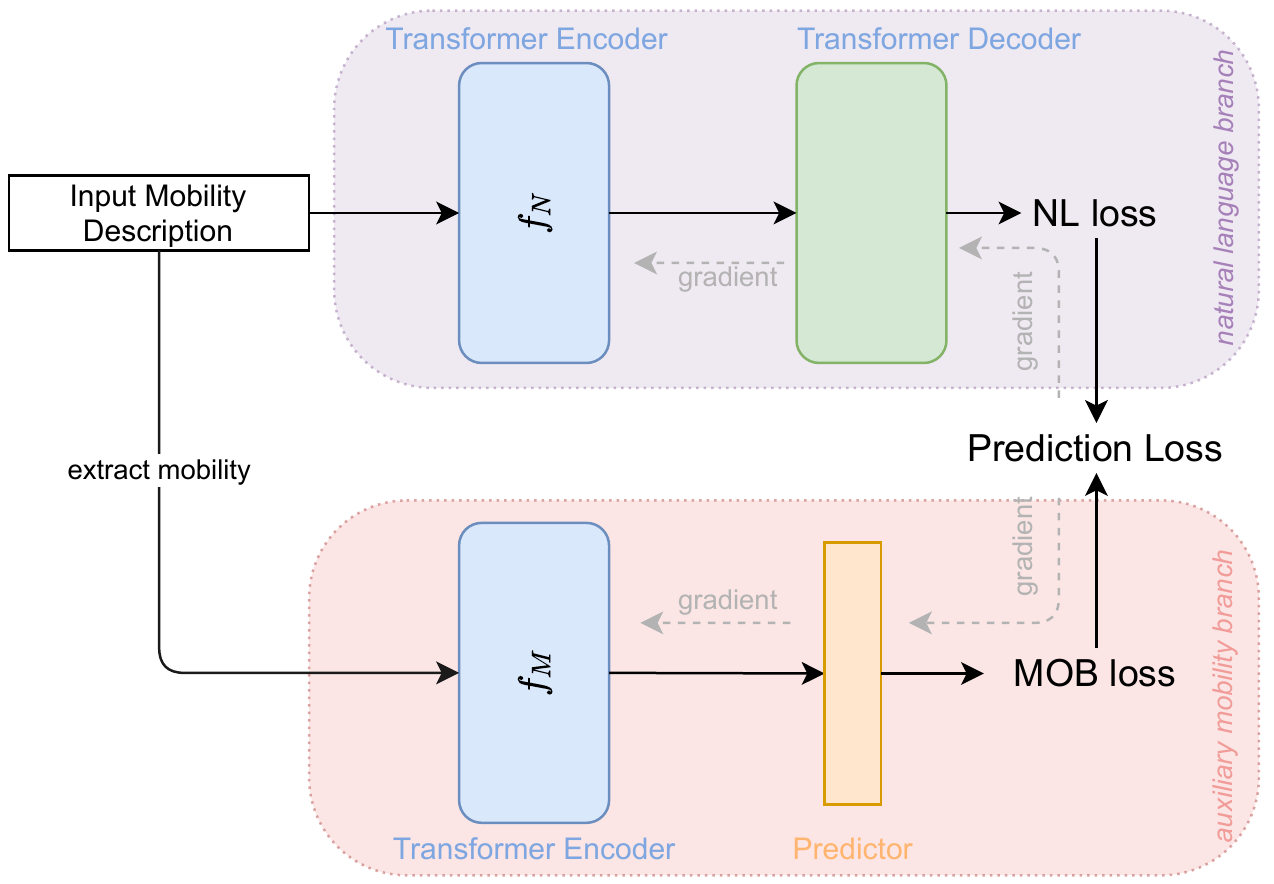}} 
    \subfigure[Siamese mode]{\label{fig:siamese}
    \includegraphics[width=.32\textwidth]{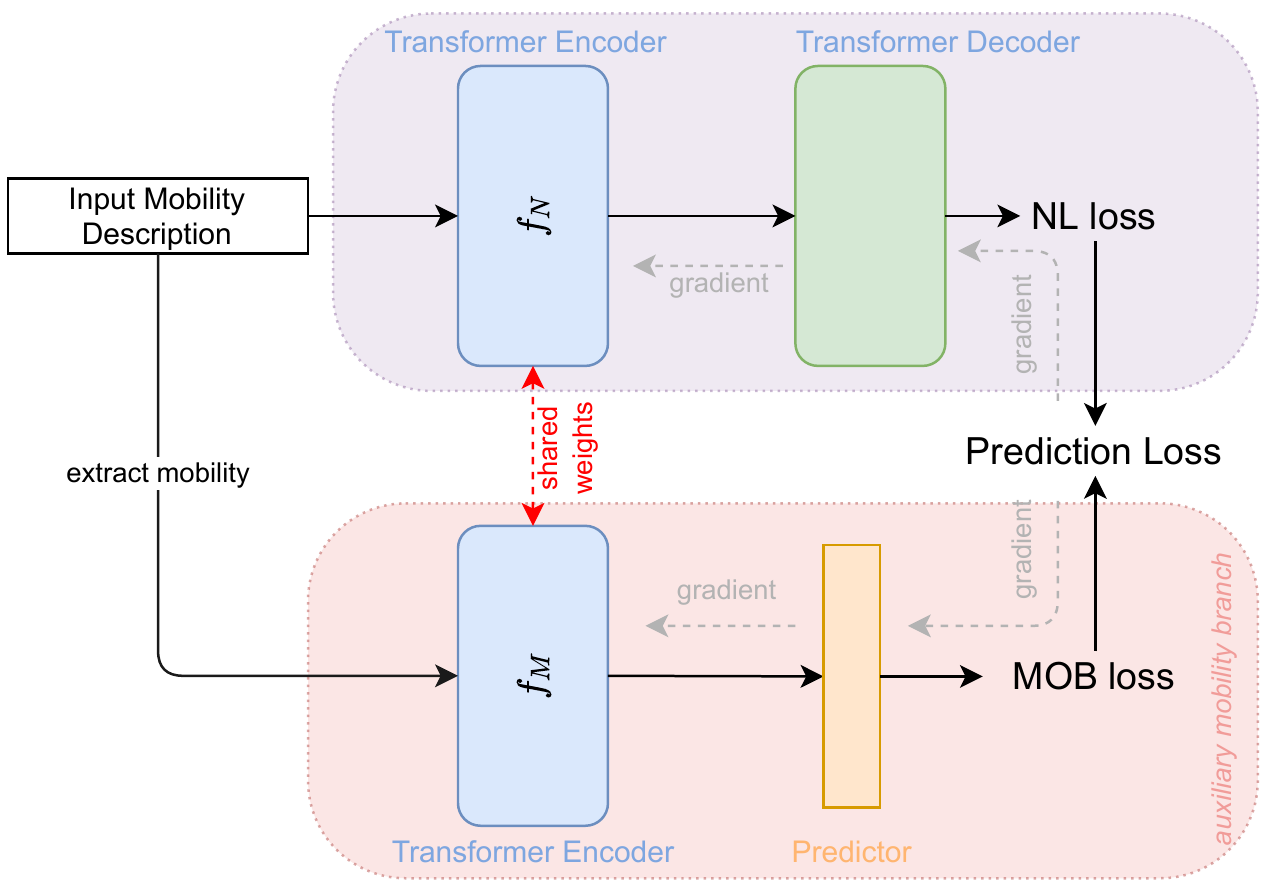}} 
    \subfigure[Momentum mode]{\label{fig:momentum}
    \includegraphics[width=.32\textwidth]{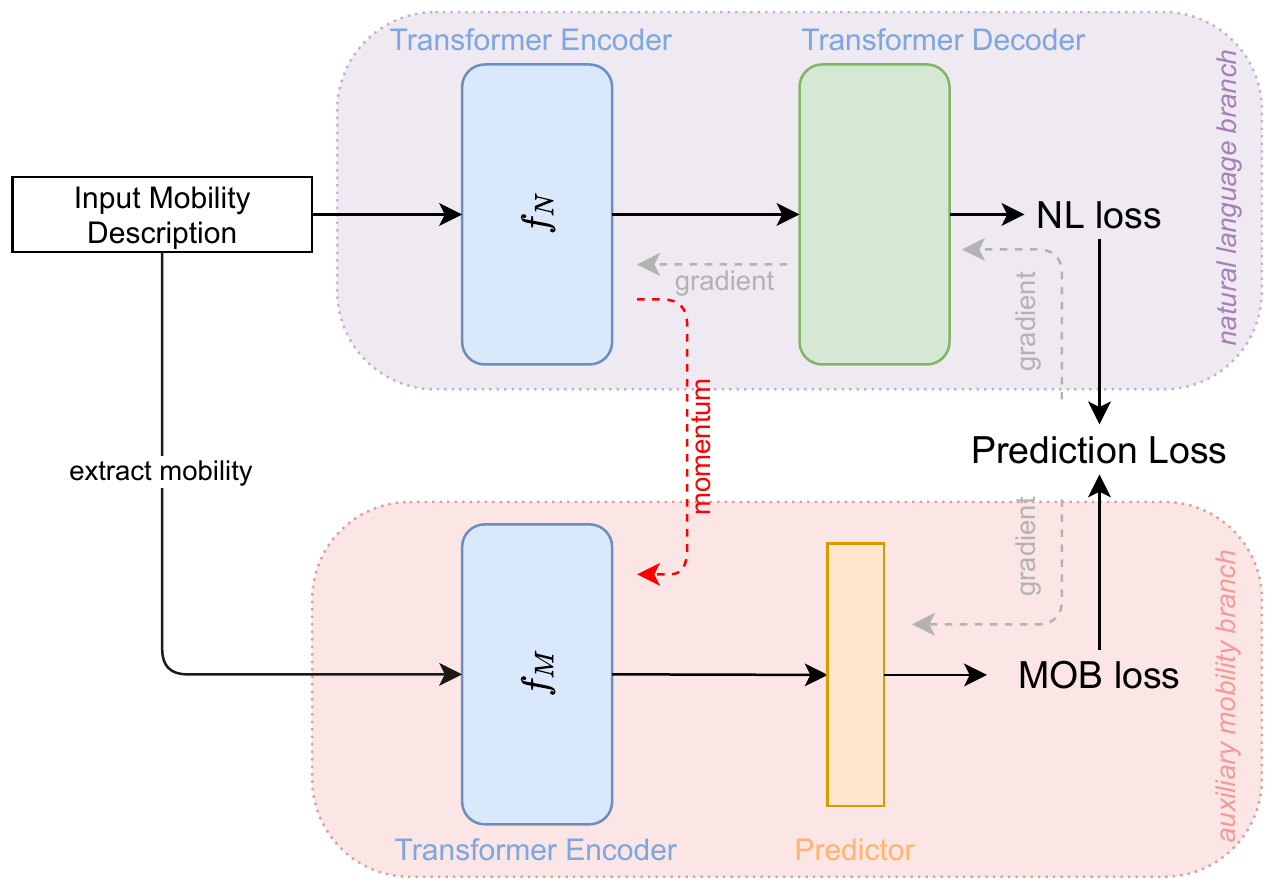}}
    \caption{Conceptual comparison of three modes of connecting two branches in~\name\ (empirical comparisons are presented in Table~\ref{tab:ablation}). In \name, the third momentum mode is selected as the default mode.}
    
\end{figure*}

\subsection{Two-Branch Structure}
The overall framework of the proposed method is illustrated in Figure~\ref{fig:momentum} (Figure~\ref{fig:basic} and Figure~\ref{fig:siamese} are two variants of our \name\ and more details are given in Section~\ref{sec:modes}).
It consists of two branches:
    (1) Natural Language Branch (\nl): a branch with the sequence-to-sequence structure, which is the main branch of \name\ to translate the input prompt to generate output sentences;
    (2) Auxiliary Mobility Branch (\mb): an auxiliary branch to strengthen the ability of \name\ in learning mobility patterns for forecasting.
The details of \name\ are given in the following sections.

\subsubsection{NL Branch}
Through mobility description, mobility data $\mathbf{X}$ and other context information (\eg, semantic category $c$) are transformed as a natural language prompt $\mathbf{S}$.
In addition, the prediction target $x_{t_{\text{obs} + 1}}$ is also described as a target sentence $Y$.
Following standard natural language processing procedures, tokenization\footnote{Tokenizer provided by HuggingFace is utilized in our implementation: https://huggingface.co/docs/tokenizers/python/latest/.} is then applied to the generated prompt sentences.

After the tokenization, the prompt $\mathbf{S}$ is interpreted as a list of tokens $[s_1, s_2, \cdots, s_J]$, where $J$ is the length of the list.
Each token (element in the list) belongs to a vocabulary where saves the token mapping of the entire dataset.
Similarly, the target sentence $Y$ (\ie, the sentence given in the last row of Table~\ref{tab:mob_language}) is encoded into $[y_1, y_2, \cdots, y_K]$ and $K$ is the length of the target sentence tokens.

The whole \nl\ branch follows the sequence-to-sequence/encoder-decoder structure and the encoding process can be formulated as:
\begin{align}
\label{eq:nl_emb}
    e_n^j &= \phi_n(s_j; \mathbf{W}_{\phi_n}), \\
    h_{N} &= f_{N} (e_n^1, e_n^2, \cdots, e_n^J; \theta_{N}), \label{eq:nl_enc}
\end{align}
where $\phi_n$ with weights $\mathbf{W}_{\phi_n}$ is an embedding layer to embed each input token into a $d$ dimension vector $e_n^j \in \mathbb{R}^d$. 
The encoder $f_{N}(\cdot)$ with trainable weights $\theta_{N}$ takes embedded vectors to yield a hidden state $h_{N}$ for the later decoder part.
In our~\name, Transformer~\cite{vaswani2017attention} is utilized as the encoder $f_{N}(\cdot)$.

The decoding part in our \nl\ branch generates predicted tokens $[\hat{y}_1, \hat{y}_2, \cdots, \hat{y}_K]$ in a autoregressive fashion.
Mathematically, the probability of decoding the $k$-th token $\hat{y}_k$ can be parameterized as:
\begin{equation}
    p(\hat{y}_k \mid \hat{y}_{<k}, h_{N}) = \text{softmax} (f_{D}(\hat{y}_{<k}, h_{N}; \theta_{D})), \label{eq:nl_pred}
\end{equation}
where $f_{D}(\cdot)$ is the decoder in the \nl\ branch.
After decoding the total $K$ tokens and applying detokenization on decoded tokens, a generated sentence $\hat{Y}$ is then obtained.

\subsubsection{Mob Branch}

Since we are particularly interested in forecasting human mobility (\eg, number of visits of each POI), an auxiliary mobility branch (\mb\ branch) is incorporated into the \name\ framework. 
As described in the above section, the \nl\ branch is a general sequence-to-sequence architecture for language generation, both mobility data related tokens (\eg, tokens represented the number of visits) and other tokens in the prompt will be treated equally.
Therefore, the motivation of introducing this auxiliary branch is to support the main \nl\ branch to better learning the mobility pattern.

For the architecture of this \mb\ branch (the lower branch in each sub-figure in Figure~2), it follows the design of typical time-series forecasting framework.
The input of this branch is the mobility data $[x_{t_1}, x_{t_2}, \cdots, x_{t_{\text{obs}}}]$ which can be extracted from the input mobility description (prompt) of the \nl\ branch or directly taken from the dataset (the raw data before mobility-to-language transformation). 

Similar to the \nl\ branch, the input of each timestamp $x_t$is first embedded into $e_m^t \in \mathbb{R}^d$ through the embedding layer $\phi_m (\cdot)$:
\begin{equation}
    e_m^t = \phi_m(x_t; \mathbf{W}_{\phi_m}) \label{eq:mb_emb}
\end{equation}
After the embedding, a Transformer-based encoder $f_{M}$ is used to extract the hidden state $h_M$:
\begin{equation}
    h_M = f_{M} (e_m^{t_1}, e_m^{t_2}, \cdots, e_m^{t_{\text{obs}}}; \theta_{M}), \label{eq:mb_enc}
\end{equation}
where $\theta_{M}$ is the weight matrix of the Transformer encoder in the \mb\ branch.
The \mb\ branch prediction ${\tilde{x}}_{t_{\text{obs}+1}}$ at time step $t_{\text{obs}+1}^i$ is then generated via:
\begin{equation}
    {\tilde{x}}_{t_{{\text{obs}}+1}} = \text{MLP}(h_M), \label{eq:mb_out}
\end{equation}
where $\text{MLP}(\cdot)$ is a multi-layer perceptrons (MLP)-based predictor.

\subsubsection{Connecting Two Branches}\label{sec:modes}
In this section, we discuss how to connect the \nl\ branch and the \mb\ branch in our \name. 
For our \name, the forecasting performance depends on the main \nl\ branch.
During the model inference phase, the \mb\ branch will be ignored as the output is in the sentence format.
As a consequence, it is more important to learn a better $f_N(\cdot)$ for the \nl\ branch.
For this purpose and inspired by~\cite{he2020momentum}, we introduce a Momentum Mode (as illustrated in Figure~\ref{fig:momentum}) to connect two encoders.
In more detail, during the training process, only $\theta_{N}$ is updated through back propagation and $\theta_{M}$ is updated via:
\begin{equation}
    \theta_{M} \leftarrow \alpha_{m}\theta_{N} + (1 - \alpha_{m}) \theta_{M} , \label{eq:momentum}
\end{equation}
where $\alpha_{m}$ is the momentum factor.
Under this mode, the \mb\ branch encoder $f_M(\cdot)$ can be seen as the momentum-based moving average of the \nl\ branch encoder $f_N(\cdot)$.
Since $\theta_{M}$ is based on $\theta_{N}$, during the training, the auxiliary \mb\ branch could support the main branch to learn a more powerful $f_N(\cdot)$ in the aspect of encoding mobility data for forecasting.

In addition to the above momentum mode, we also explore and compare the other two ways of connecting the \nl\ branch and the \mb\ branch:
    (i) Basic Mode (Figure~\ref{fig:basic}): this mode is a vanilla mode. There is no interactions between two branches except for the combined loss.
    (ii) Siamese Mode (Figure~\ref{fig:siamese}): the weights of two encoders in two branches are shared during training ($\theta_{M}=\theta_{N}$).
The comparison of using different modes is given in Section~\ref{sec:ablation}.

It is worth noting that the final prediction target can be extracted from both the \nl\ branch (${\hat{x}}_{t_{\text{obs}+1}}$ acquired from the generated sentence $\hat{Y}$) and the \mb\ branch (${\tilde{x}}_{t_{\text{obs}+1}}$ in Eq.~\eqref{eq:mb_out}).
Considering that we are interested in performing forecasting through language generation, the overall output of our \name\ is the generated sentence $\hat{Y}$ from the \nl\ branch and ${\hat{x}}_{t_{\text{obs}+1}}$ embedded in output sentence $\hat{Y}$ is used for evaluation.

\subsection{Loss Function}

As the \nl\ branch is for generating sentences, we use the conventional
multi-class cross-entropy loss function (the number of class equals to the total number of tokens in the vocabulary) given by:
\begin{equation}
    \mathcal{L}_{N} = - \sum_{b=1}^{B} \sum_{k=1}^{K} y_k^b\log \hat{y}_k^b, \label{eq:nl_loss}
\end{equation}
where $B$ is the batch size and the superscript $b$ stands for the $b$-th training sample in a batch. 
For the \mb\ branch, it is a basic time-series forecasting branch
Thus, we choose the typical mean squared error (MSE) as the loss function:
\begin{equation}
    \mathcal{L}_{M} = \frac{1}{B}\sum_{b=1}^{B} \| {\tilde{x}}_{t_{{\text{obs}}+1}} ^ b - {{x}}_{t_{{\text{obs}}+1}} ^ b \|^2. \label{eq:mb_loss}
\end{equation}

As a result, the final loss function of~\name\ is a combination of $\mathcal{L}_{N}$ and $\mathcal{L}_{M}$:
\begin{equation}
    \mathcal{L} = (1 - \alpha_{loss}) \mathcal{L}_{N} + \alpha_{loss} \mathcal{L}_{M} , \label{eq:loss}
\end{equation}
where $\alpha_{loss}$ is the loss factor to balance the two losses.
The impact of setting different $\alpha_{loss}$ is discussed in Section~\ref{sec:loss_res}.

\section{Experiments}

\subsection{Dataset}

We performed extensive experiments on real-world human mobility data presented by SafeGraph’s Weekly Patterns\footnote{https://docs.safegraph.com/docs/weekly-patterns}, which includes visitor and demographic aggregations for POIs in the US. 
It contains aggregated raw counts (no private information) of visits to POIs from a panel of mobile devices and also provides the semantic category information of each POI.
Although SafeGraph provides the data from many cities, we selected data from three major cities with different statistical features (see Table~\ref{tab:dataset} and Figure~\ref{fig:dis} in the \textit{Supplementary Material}) for building three datasets: New York City (NYC), Dallas, and Miami.
Since some POIs only have visiting records for several weeks,
we first filter out POIs without complete visiting records during the entire data collection period.
The mobility-to-language template introduced in Section~\ref{sec:des} is then applied to generate natural language sentences to form datasets.
Each dataset is randomly divided into the training set (70\%), validation set (10\%), and testing set (20\%).
\begin{table}[]
\centering
\caption{Details of three datasets.}
\begin{tabular}{l|ccc} \hline
                       & NYC & Dallas & Miami  \\ \hline
Collection Start Date  & \multicolumn{3}{c}{2020-06-15}  \\
Collection End Date    & \multicolumn{3}{c}{2020-11-08}  \\
Average Visits per Day & 17.082        & 21.520  & 22.977 \\
Max Number of Visits   & 246           & 2746   & 1550   \\
Total Number of POIs    & 479           & 1374   & 1007   \\
Number of Categories     & 39            & 65     & 51     \\ \hline
\end{tabular}
\label{tab:dataset}
\end{table}
Table~\ref{tab:dataset} shows the statistics (after filtering) of the datasets.
Based on the table, it can be seen that three selected datasets have different levels in the total number of POIs, max number of visits, and the number of semantic categories.
This ensures the representativeness of our data used for experiments.

\begin{table*}[]
\centering
\caption{Performance results of different methods on the three datasets. In each row, the best performer is shown in bold and the second best is given in blue.}
\addtolength{\tabcolsep}{-0.35ex}
\begin{tabular}{l||cc|cc|cc|cc} \hline
 & \multicolumn{2}{c|}{NYC} & \multicolumn{2}{c|}{Dallas} & \multicolumn{2}{c|}{Miami} & \multicolumn{2}{c}{Average} \\ \cline{2-9}
 & RMSE & MAE & RMSE & MAE & RMSE & MAE & RMSE & MAE \\ \hline
LR & 9.131 & 5.639 & 24.544 & 6.601 & 13.081 & 6.082 & 15.585 & 6.107\\
Gru & 7.547 (0.098) & 4.550 (0.038) & 23.987 (0.262) & 5.400 (0.016) & 12.125 (0.160) & 5.413 (0.026) & 14.553 & 5.121 \\
GruA & 7.704 (0.107) & 4.464 (0.037) & 22.562 (0.433) & 5.276 (0.048) & 11.465 (0.417) & 5.045 (0.107) & 13.910 & 4.928 \\
Transformer & 6.714 (0.072) & 4.279 (0.058) & 18.820 (0.278) & 5.166 (0.125) & 10.995 (0.181) & 5.130 (0.117) & 12.176 & 4.858 \\
Reformer & 6.626 (0.061) & 4.395 (0.074) & \textcolor{blue}{17.392 (0.178)} & 5.120 (0.037) & 10.578 (0.242) & 5.117 (0.065) & 11.532 & 4.877 \\
Informer & \textcolor{blue}{6.509 (0.073)} & \textbf{4.248 (0.065)} & 19.386 (0.383) & 6.717 (0.453) & 9.858 (0.171) & 5.159 (0.103) & 11.918 & 5.375 \\ \hline
\pname(GruA) & 6.901 (0.212) & 4.290 (0.042) & 19.914 (1.259) & 5.165 (0.067) & 9.964 (0.632) & 5.009 (0.055) & 12.260 & 4.821 \\
\pname(Transformer) & 6.657 (0.070) & 4.286 (0.075) & 18.212 (1.422) & 5.036 (0.096) & \textcolor{blue}{9.672 (0.605)} & 5.034 (0.105) & \textcolor{blue}{11.514} & 4.785 \\
\pname(BART) & 6.645 (0.166) & 4.313 (0.232) & 18.978 (2.102) & \textcolor{blue}{4.968 (0.045)} & 9.724 (0.307) & \textbf{4.834 (0.016)} & 11.782 & \textbf{4.705} \\ \hline
\name\ & \textbf{6.426 (0.067)} & \textcolor{blue}{4.274 (0.049)} & \textbf{15.248 (0.367)} & \textbf{4.928 (0.043)} & \textbf{8.580 (0.159)} & \textcolor{blue}{4.951 (0.028)} & \textbf{10.085} & \textcolor{blue}{4.718} \\ \hline
\end{tabular}
\label{tab:results}
\end{table*}

\subsection{Implementation Details}

The hidden dimension $d$ for the Transformer is chosen as 256 for both the main \nl\ branch and the auxiliary \mb\ branch.
To avoid over-fitting, the dropout rate is set as 0.2.
The hyperparameters are set based on the performance of the validation set.
The total number of training epochs is 36 with batch size 128 (for the Dallas and Miami) or batch size 64 (for the NYC).
The loss factor $\alpha_{loss}$ and the momentum factor $\alpha_{m}$ are selected as 0.01 and 0.001, respectively.
The proposed methods are optimized with Adam optimizer~\cite{kingma2014adam} (a 0.0001 initial learning rate with \textit{ReduceLROnPlateau} decay) on a desktop with an NVIDIA GeForce RTX-2080 Ti GPU with PyTorch.

\subsection{Prediction Performance}

\subsubsection{Baselines for Comparison}

As comparison, we select 9 methods which are classified into two different categories:
\begin{itemize}
    \item Time-series forecasting methods:
    (1) basic linear regression (LR);
    (2) Gru~\cite{chung2014empirical}: Gated Recurrent Units, one of basic RNNs;
    (3) GruA~\cite{BahdanauCB2015}: Gru with attention mechanism; 
    (4) Transformer \cite{vaswani2017attention}: the vanilla Transformer structure. This can considered as a model only using the \mb\ branch of our~\name.
    (5) Reformer~\cite{reformer2020}: an efficient variant of Transformer;
    (6) Informer \cite{zhou2021informer}: a state-of-the-art Transformer variant specifically designed for time-series prediction.
    \item Natural language sequence-to-sequence structure (\pname): (1) using GruA network as the backbone;
    (2) using Transformer network as the backbone: this can considered as a model only using the \nl\ branch of our~\name.
    (3) BART~\cite{lewis2020bart}: a recent Transformer-based architecture which has a bidirectional encoder and an autoregressive decoder. It is designed for natural language sequence-to-sequence tasks. Note that the pre-trained weights of this network is not used in the experiments for fair comparison.\footnote{The configuration of this model can be accessed through: https://huggingface.co/facebook/bart-base/tree/main.}
\end{itemize}
For the first category methods, the typical time-series forecasting framework (the upper one in Figure~\ref{fig:intro}) is applied.
The proposed forecasting through language generation pipeline (the lower one in Figure~\ref{fig:intro}) is utilized for the methods under the second category.

\begin{table*}[]
\centering
\caption{The prediction results of the ablation studies on all three datasets, without (w/o) the specifically mentioned branches or using different modes.}
\addtolength{\tabcolsep}{-0.5ex}
\begin{tabular}{l||cc|cc|cc|cc} \hline
 & \multicolumn{2}{c|}{NYC} & \multicolumn{2}{c|}{Dallas} & \multicolumn{2}{c|}{Miami} & \multicolumn{2}{c}{Average} \\ \cline{2-9}
 & RMSE & MAE & RMSE & MAE & RMSE & MAE & RMSE & MAE \\ \hline
\name~(w/o NL branch) & 6.714 (0.072) & 4.279 (0.058) & 18.820 (0.278) & 5.166 (0.125) & 10.995 (0.181) & 5.130 (0.117) & 12.176 & 4.858 \\
\name~(w/o Mob branch) & 6.657 (0.070) & 4.286 (0.075) & 18.212 (1.422) & 5.036 (0.096) & 9.672 (0.605) & 5.034 (0.105) & 11.514 & 4.785 \\ \hline
\name~(basic) & 6.493 (0.056) & 4.365 (0.057) & 18.292 (1.715) & 5.037 (0.124) & 8.792 (0.254) & 5.133 (0.093) & 11.192 & 4.845 \\
\name~(siamese) & 6.519 (0.077) & 4.317 (0.014) & 19.311 (1.242) & 5.212 (0.091) & 8.760 (0.155) & 5.020 (0.068) & 11.530 & 4.850 \\ \hline
\name\ & 6.426 (0.067) & 4.274 (0.049) & 15.248 (0.367) & 4.928 (0.043) & 8.580 (0.159) & 4.951 (0.028) & 10.085 & 4.718
\\ \hline
\end{tabular}
\label{tab:ablation}
\end{table*}

\subsubsection{Evaluation Protocol and Metrics}
To evaluate the performance of different methods, we report
two widely used metrics for prediction tasks: the Root Mean Square Error (RMSE) and the Mean Absolute Error (MAE).
For the proposed~\name\ and other \pname-based methods, the direct outputs are sentences (\eg, the last row of Table~\ref{tab:mob_language}). 
Thus, $\hat{x}_{t_{\text{obs} + 1}}$ is firstly decoded from each output sentence before calculating RMSE and MAE.
In the following experiments, we report the average performance (and the standard deviation) of 5 runnings of each method (excluding LR) or configuration.

\subsubsection{Performance}
Table~\ref{tab:results} shows the results of different methods. 
The average performance across all three datasets is also given in the last two columns of the table.
In general, we observe that \name\ consistently outperforms all baseline techniques in RMSE (12.4\% performance gain, compared to the second best) and achieves the second best in average MAE (only about 0.2\% worse than the best performer BART). 
Compared to other methods, \name\ brings a significant RMSE improvement especially on the Dallas and Miami datasets which are more difficult (due to more POIs and larger range of the number of visits value) to predict.
For the MAE metric, our \name\ is the top performer on Dallas and other top performers are Informer and \pname(BART).
Note that although \pname(BART) slightly outperforms our \name\ on average MAE, the computational cost of \pname(BART) is significantly larger than \name\ (see Table~\ref{tab:cost} given in the \textit{Supplementary Material}).
These results demonstrate the effectiveness of the proposed \name.

In addition, if we compare methods using the same network architecture, \pname(GruA) leads GruA with an improvement of 11.9\% in RMSE and \pname(Transformer) outperforms Transformer by around 5.4\% in RMSE.
It can be seen that applying the proposed forecasting through language generation pipeline (\pname) is able to boost human mobility forecasting performance and \pname\ is robust to work with different prediction neural network architectures.

\subsection{Ablation Study} \label{sec:ablation}
In this part, we conducted experiments on three datasets with ablation consideration.
To evaluate each branch and different connecting modes of \name, the following variants are compared:
\begin{itemize}
    \item Without the \nl\ branch: the \nl\ branch of \name\ is disabled. This variant is the same as Transformer in Table~\ref{tab:results}.
    \item Without the \mb\ branch: the \mb\ branch is removed. This variant is the same as \pname(Transformer) in Table~\ref{tab:results}.
    \item \name\ (basic): using Basic mode to connect two branches (details given in Section~\ref{sec:modes}).
    \item \name\ (siamese): using Siamese mode to connect two branches (details given in Section~\ref{sec:modes}).
\end{itemize}

The results of these variants and our \name\ (using the default momentum mode) on the three datasets are given in Table~\ref{tab:ablation}.
From the table, we can observe that: (1) The proposed \name\ greatly
outperforms the first two variants where only one branch is enabled. 
It justifies the need of incorporating both branches.
(2) The momentum mode shows better performance than the basic and the siamese modes. Specifically, using the siamese mode has the worst performance. It is even worse than \name~(w/o Mob branch), which indicates that the auxiliary \mb\ branch has a negative effect on prediction performance under the siamese mode setting.
Based on the above results, we conclude that the momentum mode is more suitable for the proposed two-branch \name.

\begin{table*}[]
\caption{The prediction performance of using different Prompts in \name.}
\addtolength{\tabcolsep}{-0.15ex}
\begin{tabular}{l||cc|cc|cc|cc} \hline
 & \multicolumn{2}{c|}{NYC} & \multicolumn{2}{c|}{Dallas} & \multicolumn{2}{c|}{Miami} & \multicolumn{2}{c}{Average} \\ \cline{2-9}
 & RMSE & MAE & RMSE & MAE & RMSE & MAE & RMSE & MAE \\ \hline
Prompt A & 6.426 (0.067) & 4.274 (0.049) & 15.248 (0.367) & 4.928 (0.043) & 8.580 (0.159) & 4.951 (0.028) & 10.085 & 4.718 \\
Prompt B (w/o semantic) & 6.496 (0.091) & 4.301 (0.047) & 16.035 (0.633) & 5.032 (0.012) & 8.688 (0.223) & 4.982 (0.041) & 10.406 & 4.772 \\ \hline
\end{tabular}
\label{tab:prompt}
\end{table*}

\begin{figure*}[]
    \centering
    \subfigure[Loss Factor: RMSE]{\label{fig:diff_loss_rmse}
    \includegraphics[width=.23\textwidth]{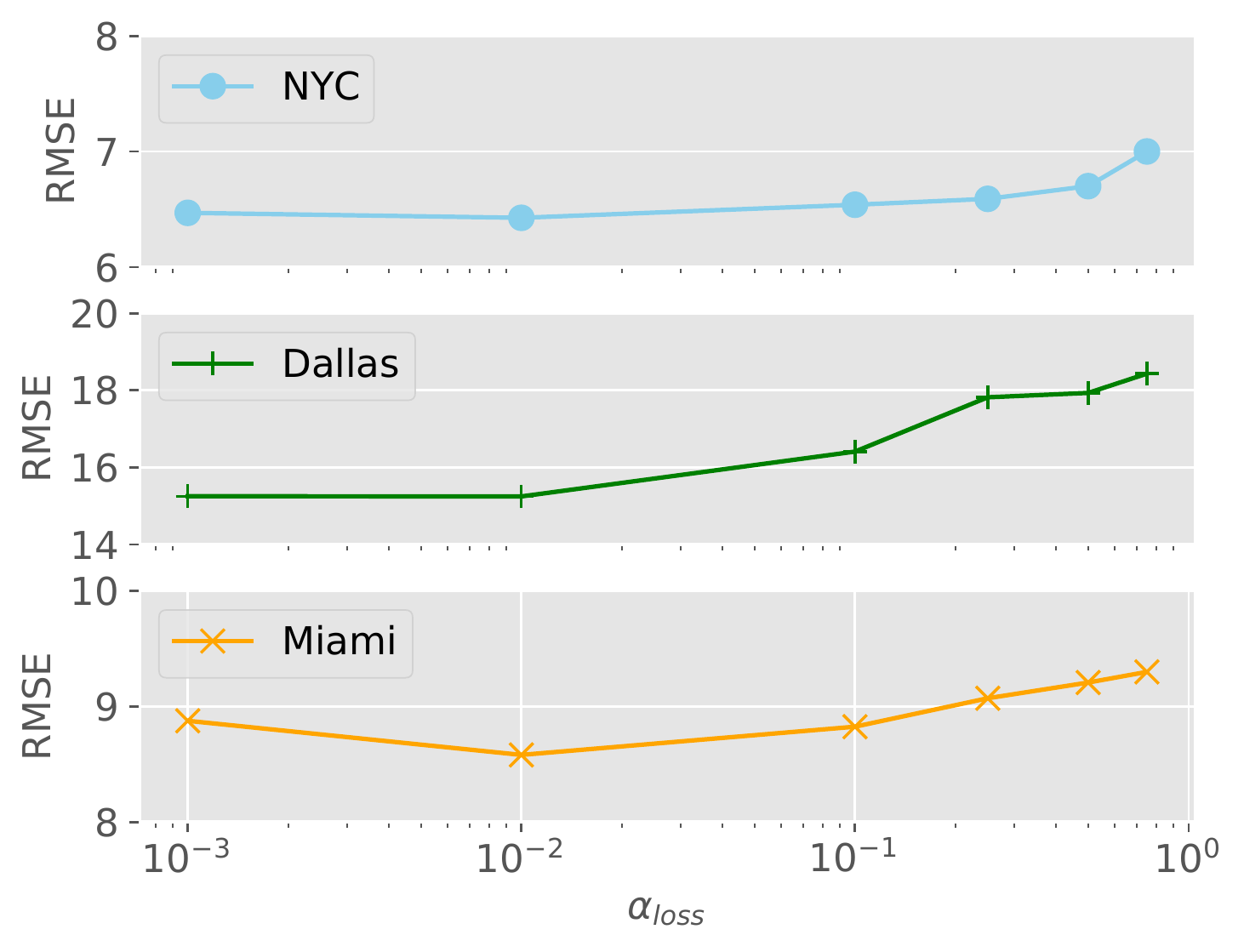}} 
    \subfigure[Loss Factor: MAE]{\label{fig:diff_loss_mae}
    \includegraphics[width=.23\textwidth]{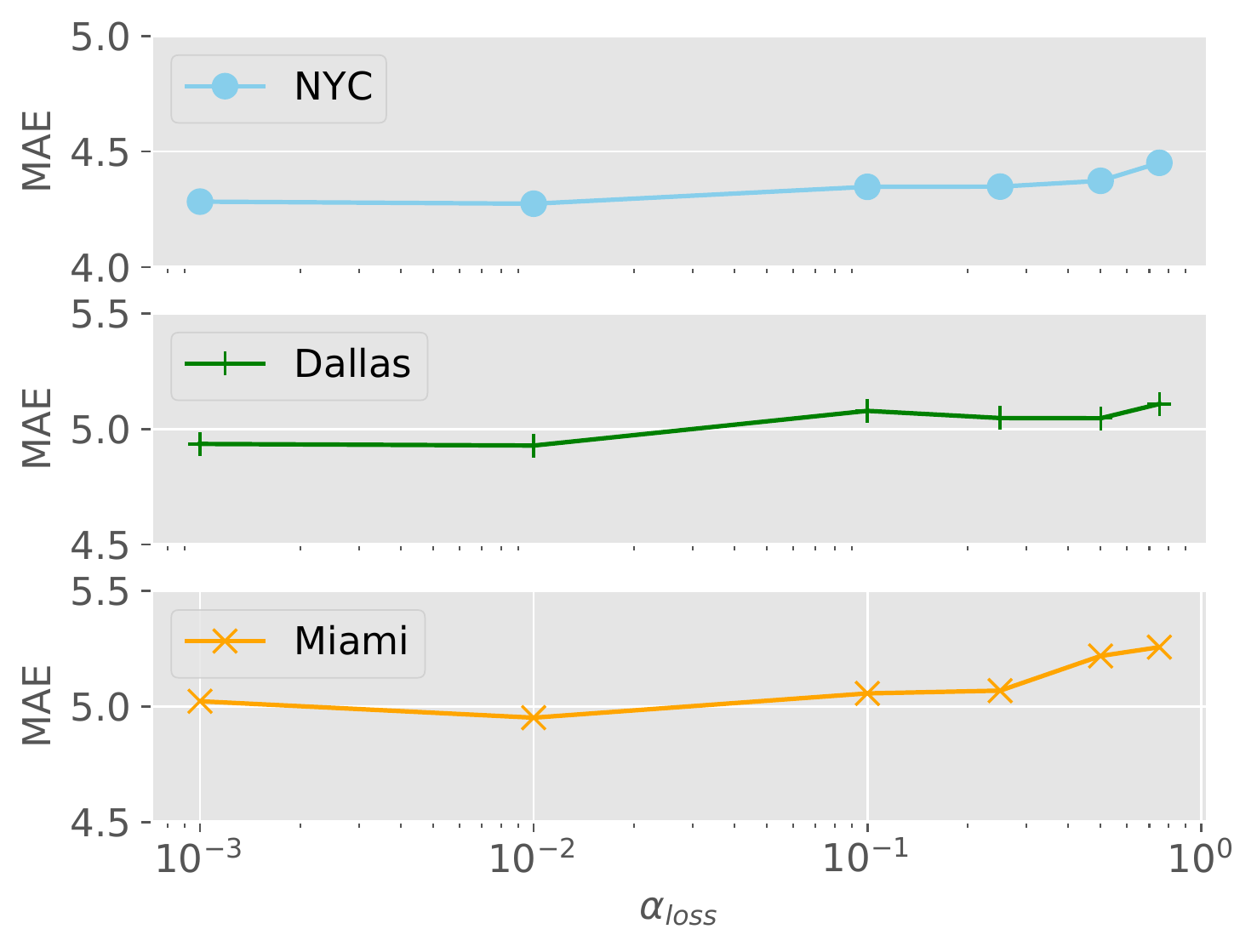}} 
    \subfigure[Momentum Factor: RMSE]{\label{fig:diff_m_rmse}
    \includegraphics[width=.23\textwidth]{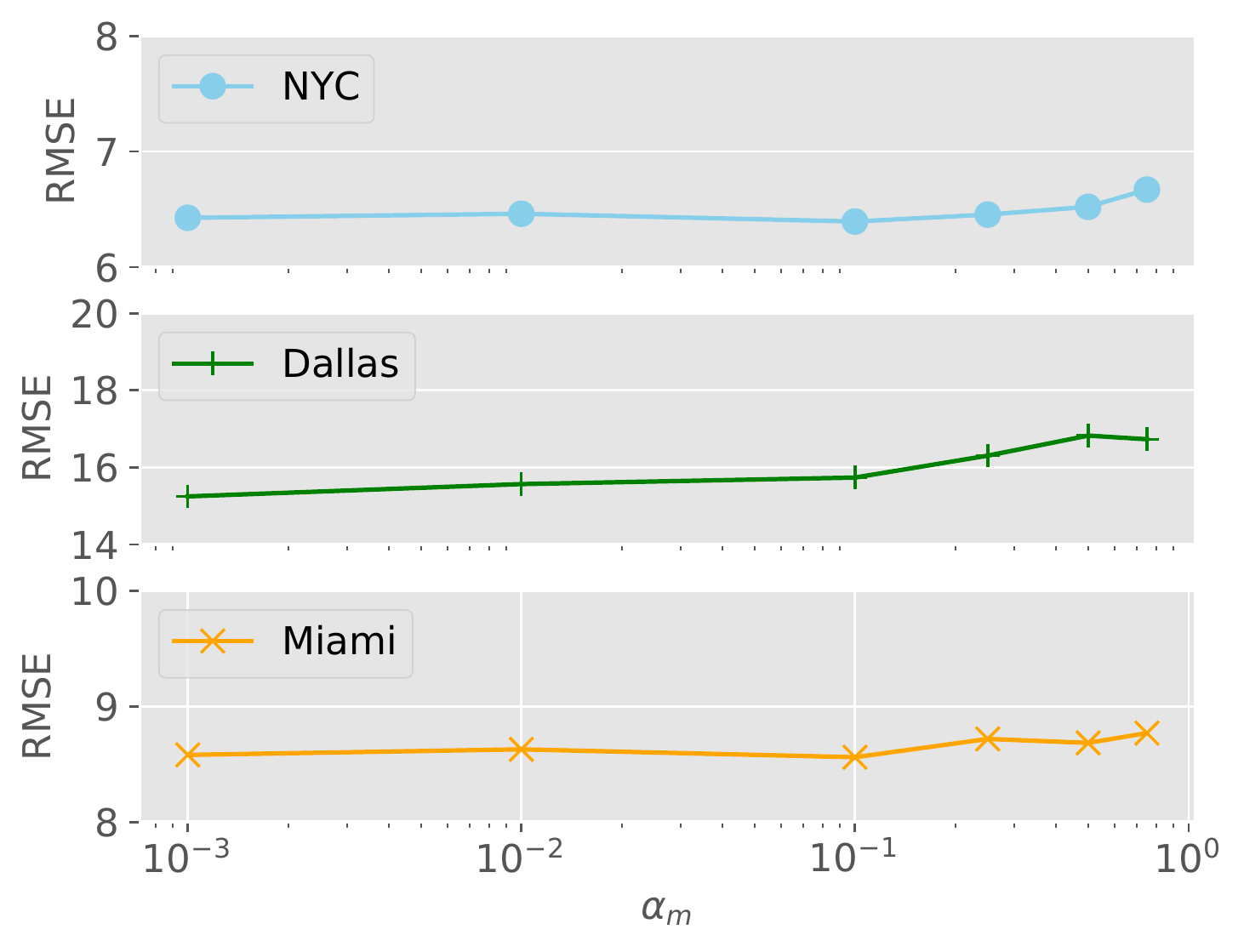}} 
    \subfigure[Momentum Factor: MAE]{\label{fig:diff_m_mae}
    \includegraphics[width=.23\textwidth]{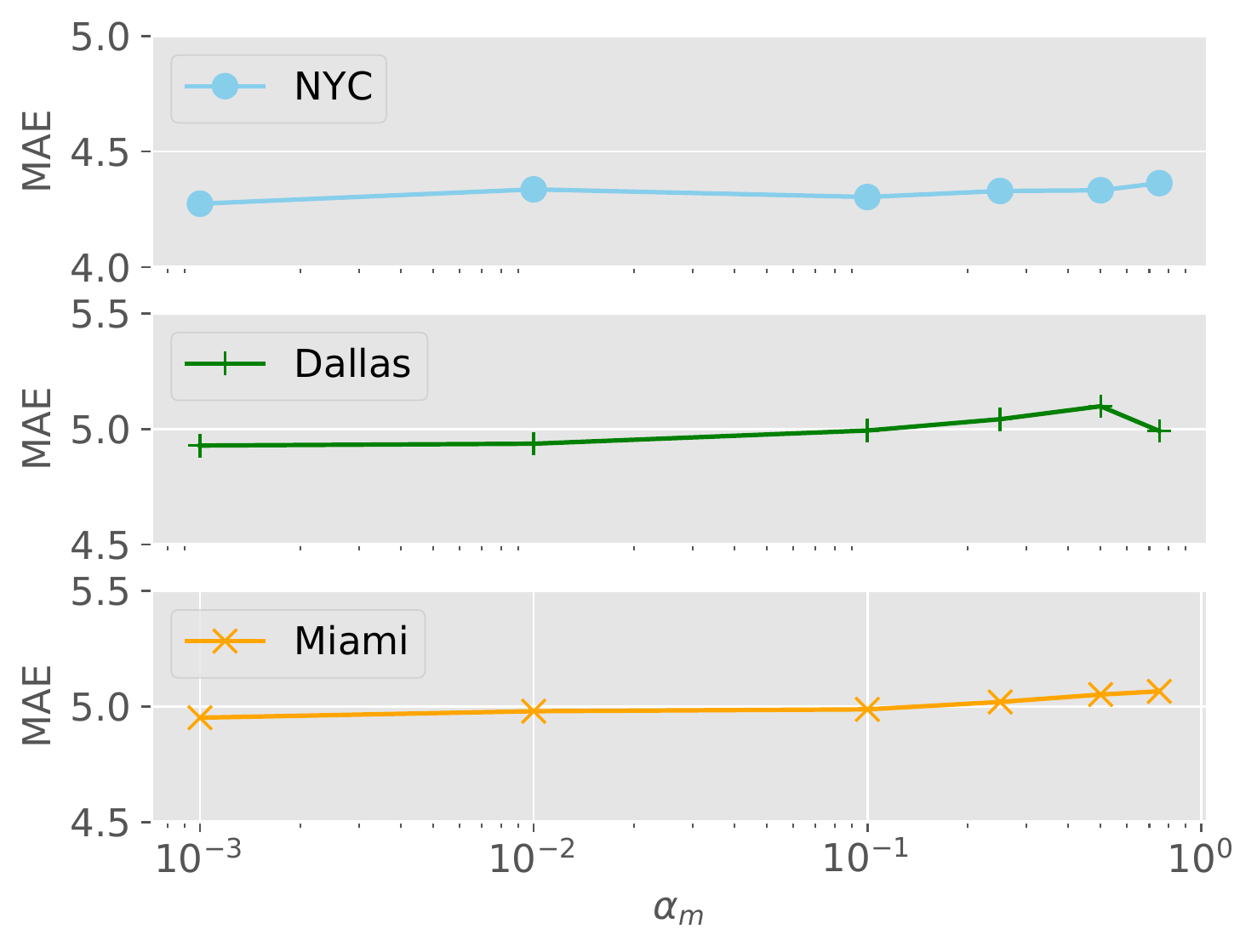}} 
    \caption{The impact of loss factor $\alpha_{loss}$ and momentum factor $\alpha_{m}$.}
\end{figure*}

\subsection{Impact of Different Settings}
\subsubsection{Different Prompts}\label{sec:prompt}
In the proposed forecasting via language generation pipeline, the mobility description is an important factor. 
We explore the impact of different prompts on mobility forecasting performance.
To be specific, two types of prompts are used as the input of our \name:
\begin{itemize}
    \item Prompt A: This is the default prompt for the proposed method. It contains all the elements listed in Table~\ref{tab:mob_language}.
    \item Prompt B: Compared to Prompt A, the sentence used for describing the POI semantic category information (the first row in Table~\ref{tab:mob_language}) is removed in Prompt B.
\end{itemize}
As given in Table~\ref{tab:prompt}, using Prompt A consistently produces performance improvements over using Prompt B on all three datasets. 
This indicates that incorporating external semantic information of POIs is also beneficial to the mobility prediction task under the forecasting via language generation pipeline.


\subsubsection{Different Loss Factors}\label{sec:loss_res}
In this experiment, we analysis the impact of the loss factor $\alpha_{loss}$ on the performance of \name\ by varying $\alpha_{loss}$ from \{0.001, 0.01, 0.1, 0.25, 0.5, 0.75\}.
The average (of 5 runnings) RMSE and MAE of \name\ with different $\alpha_{loss}$ settings on all three datasets are shown in Figure~\ref{fig:diff_loss_rmse} and \ref{fig:diff_loss_mae}, respectively.
We can observe that a smaller $\alpha_{loss}$ ($< 0.1$) leads to a well performance. When a larger $\alpha_{loss}$ is applied, the prediction performance of \name\ drops considerably.
During the training of \name, it can be noticed that $\mathcal{L}_M$ (MSE loss) has a relatively larger value than $\mathcal{L}_N$ (cross-entropy loss).
Thus, a smaller $\alpha_{loss}$ could better balance these two loss terms, which results in a better prediction performance. 

\subsubsection{Different Momentum Factors}
In this part, we investigate the impact of the momentum factor by selecting
$\alpha_{m}$ from \{0.001, 0.01, 0.1, 0.25, 0.5, 0.75\}.
The average RMSE and MAE of 5 runnings using different $\alpha_{m}$ setting is given in  Figure~\ref{fig:diff_m_rmse} and \ref{fig:diff_m_mae}.
From these two figures, it can be seen that \name\ performs reasonably well $\alpha_{m}$ is small (ranging from 0.001 to 0.1).
When the $\alpha_{m}$ becomes larger, we observe that there is an increasing RMSE for NYC and Dallas, whereas the performance is relatively stable with all $\alpha_{m}$ values for the Miami dataset.
From these results, although a smaller $\alpha_{m}$ (a slowly updating $f_M$ for the \mb\ branch) is beneficial, the impact of the momentum factor $\alpha_{m}$ is less than the loss factor $\alpha_{loss}$.

\subsubsection{Different Observation Lengths}

\begin{figure}[]
    \centering
    \subfigure[RMSE]{\label{fig:diff_len_rmse}
    \includegraphics[width=.22\textwidth]{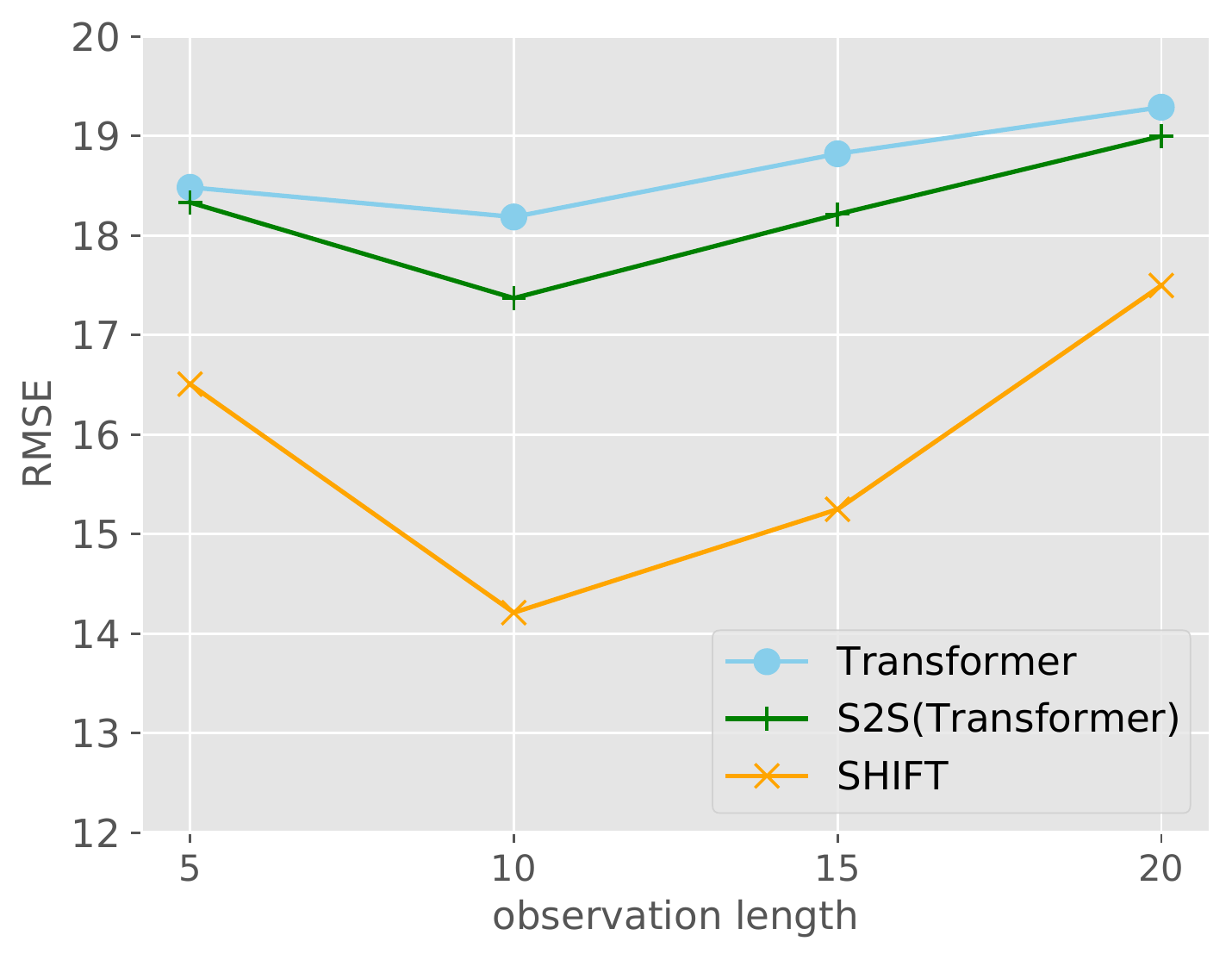}} 
    \subfigure[MAE]{\label{fig:diff_len_mae}
    \includegraphics[width=.22\textwidth]{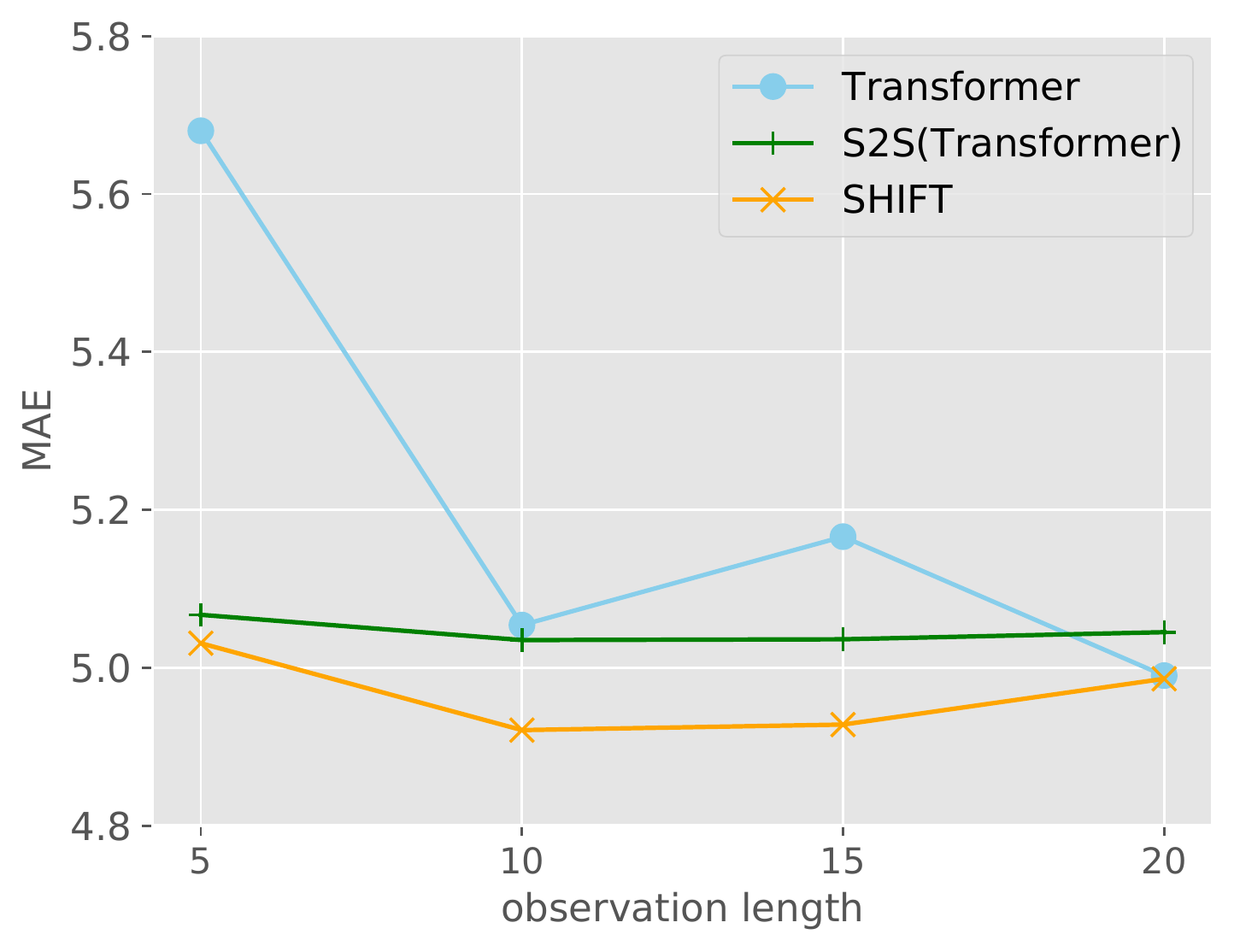}} 
    \caption{The performance of different observation lengths.}
\end{figure}

\begin{figure*}[]
    \centering
    \subfigure{
    \includegraphics[width=.95\textwidth]{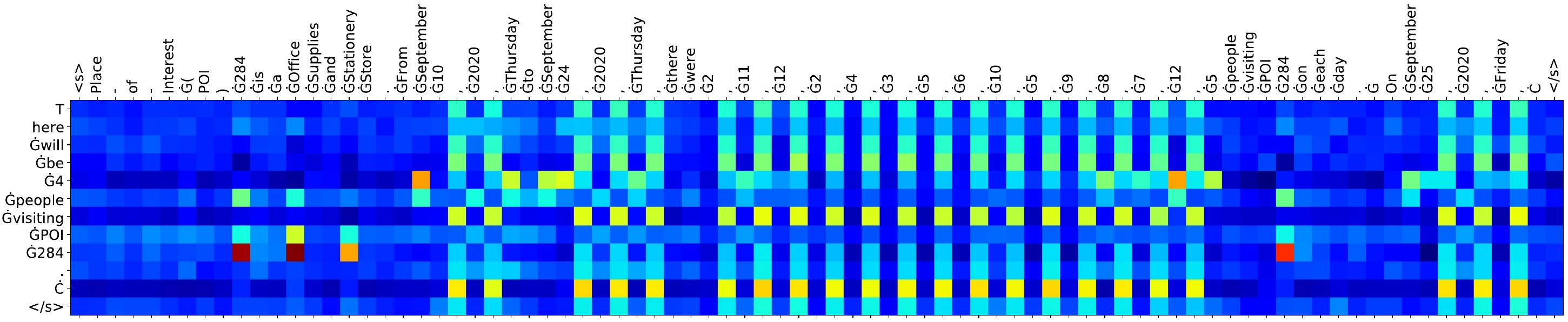}} \\ \vspace{-3ex}
    \subfigure{
    \includegraphics[trim={0 0 0 1.75cm},clip, width=.95\textwidth]{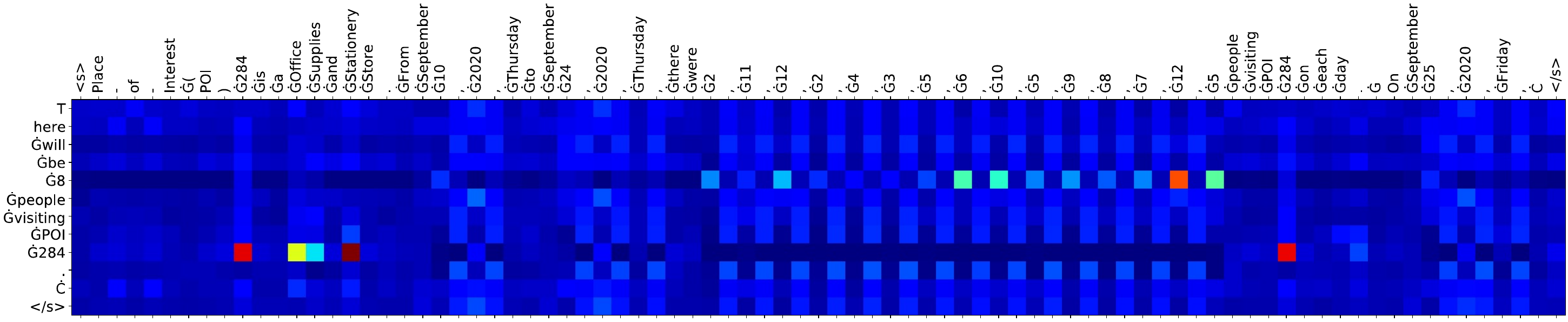}} 
    \subfigure{
    \includegraphics[width=.95\textwidth]{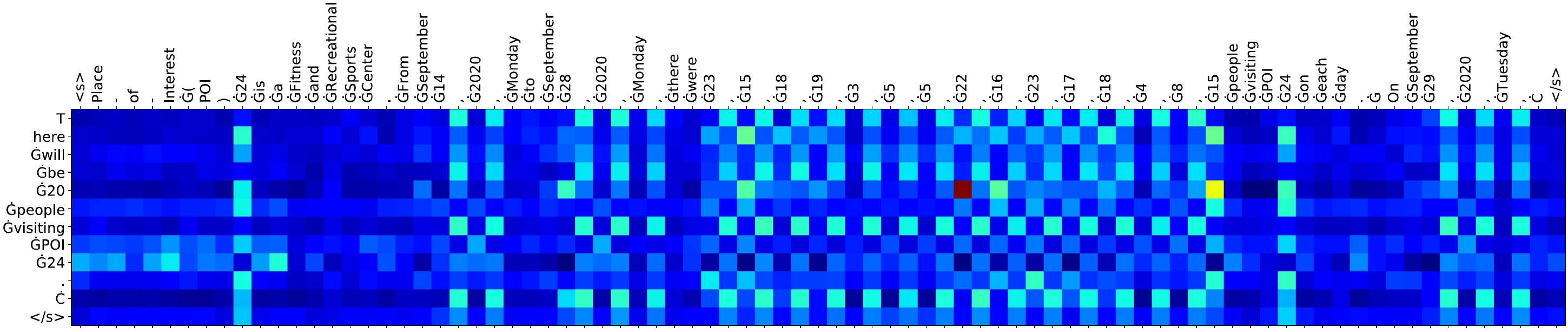}} \\ \vspace{-3ex}
    \subfigure{
    \includegraphics[trim={0 0 0 1.95cm},clip, width=.95\textwidth]{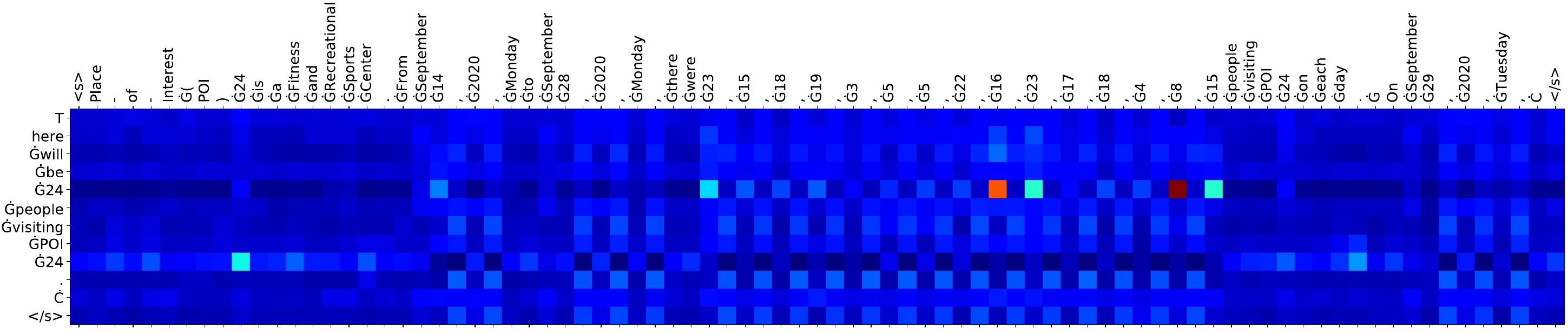}} 
    \caption{Visualization of two cases (one in the upper half and one in the lower half). For each case, the first row shows the attention of \pname(Transformer) and the attention of \name\ is presented in the second row.  (Better viewed in color.)}
    \label{fig:vis}
\end{figure*}

In the last part, we evaluate the performance of \name\ with different observation lengths. Specifically, we compare the performance of Transformer, \pname(Transformer), and \name\ with the observation length as 5, 10, 15, 20, respectively.
Due to the large amount of experiments (3 methods and 4 different observation lengths), we only report the performance (average of 5 runnings) using the challenging Dallas dataset in Figure~\ref{fig:diff_len_rmse} (RMSE) Figure~\ref{fig:diff_len_mae} (MAE).
From the figure, it can be seen that our \name\ outperforms both Transform and \pname(Transformer) on all different observation lengths.
Such an observation demonstrates the superior prediction performance of \name. 
We can also observe that all three methods have relatively worse results when the observation length is too small (${\text{obs}}=5$) or too large (${\text{obs}}=20$).
A smaller observation length leads to fewer available history records to learn mobility patterns, whereas a very large observation length might also increase the difficulty to discover effective mobility patterns for prediction.
It suggests that a favorable observation length for \name\ is roughly one or two weeks.

\subsection{Visualization Analysis}

In Figure~\ref{fig:vis}, we visualize the attentions (between the input sentence and the output sentence) learned by the Transformer encoder-decoder architecture in \name. The upper half of Figure~\ref{fig:vis} is the first case and the lower half gives the second case.
In each half (each case), the first row is the learned attentions of \pname(Transformer) (\name\ without the \mb\ branch), whereas the second row illustrates the learned attentions of our \name\ (both \nl\ and \mb\ branches).
For each heatmap plot in which a hotter region means a larger attention value, the horizontal axis stands for the input prompt (in the token format) and the vertical axis represents the output sentence tokens.
In more detail, {\fontfamily{qcr}\selectfont <s>}, {\fontfamily{qcr}\selectfont</s>}, {\fontfamily{qcr}\selectfont $\dot{\text{C}}$} are the sentence starting token, sentence ending token, and padding token, respectively.

\subsubsection{Case Analysis 1}

The ground truth label of this case is: {\fontfamily{qcr}\selectfont There will be 9 people visiting POI 284.} From the upper half of the figure, it can be seen that \pname(Transformer) generates {\fontfamily{qcr}\selectfont There will be 4 people visiting POI 284.} and the \name\ predicts the number of visits as 8. 
As a comparison, the prediction of only using the \mb\ branch (Transformer method in Table~\ref{tab:results}, not shown in the figure) is $7.71$.
From the visualization of \pname(Transformer) and \name, we observe that: (1) When generating the POI id (\eg, $284$) in the output sentence, both models would look into not only the POI id given in the prompt but also the semantic information (\eg, higher attention values on the {\fontfamily{qcr}\selectfont Office} and {\fontfamily{qcr}\selectfont Stationery} tokens).
(2) The attentions of \pname(Transformer) has an evener distribution across the input and output tokens, whereas the attentions learned by \name\ focus more on the mobility data (the {\fontfamily{qcr}\selectfont $\dot{\text{G}}$8} row).
Due to the \mb\ branch, \name\ would particularly learn mobility patterns from the mobility data tokens (\ie, the number of visits values in the input sentence), which leads to high attentions on the mobility tokens and a better prediction performance. This result supports our motivation of introducing the \mb\ branch.

\subsubsection{Case Analysis 2}
The ground truth label of this case is: {\fontfamily{qcr}\selectfont There will be 24 people visiting POI 24.} and the Transformer method yields a prediction of $18.41$ for this example.
As the POI id and the predicted number of visits are the same value (both $24$), this case is more difficult.
It requires the model to distinguish the same number with different meanings.
From the first row of the second case (lower half of Figure~\ref{fig:vis}), it can be noticed that the {\fontfamily{qcr}\selectfont $\dot{\text{G}}$20} row (predicted mobility) has a relatively high attention on the POI id in the input tokens ({\fontfamily{qcr}\selectfont $\dot{\text{G}}$24} columns), while the {\fontfamily{qcr}\selectfont $\dot{\text{G}}$24} row (predicted POI id) has very low attention values on the input POI id tokens.
It indicates that the \pname(Transformer) cannot well distinguish the mobility and the POI id in this case. This further results in worse prediction performance.
However, with the help of the auxiliary \mb\ branch, \name\ still can recognize and concentrate on the number of visits (the mobility data part) for this challenging case (the last row of Figure~\ref{fig:vis}). This demonstrates the superior prediction performance of the proposed \name.

\section{Conclusion}
We address the human mobility forecasting problem in a natural language translation manner.
Based on our mobility description template, mobility data is transformed into natural language sentences.
Through a sequence-to-sequence mobility translation, a sentence indicating the predicted mobility is then generated as output.
Furthermore, we have designed a two-branch \name\ architecture to perform mobility translation.
Through extensive experiments, the results illustrate that \name\ is effective for the human mobility forecasting task.
The effectiveness of each branch and the momentum mode in \name\ are also demonstrated in ablation studies.
In the future, we will focus on developing a method for automatic mobility-to-language generation, which aims at providing diverse prompts for the language generation-based mobility prediction.

\begin{acks}
We would like to acknowledge the support Australian Research Council (ARC) Discovery Project \textit{DP190101485}.
We thank SafeGraph for providing free access to the mobility data.
\end{acks}

\bibliographystyle{ACM-Reference-Format}
\bibliography{main}


\clearpage 
\appendix

\section{Supplementary Material}

\subsection{Dataset Distribution}
\begin{figure}[]
    \centering
    \subfigure[NYC]{
    \includegraphics[width=.3\textwidth]{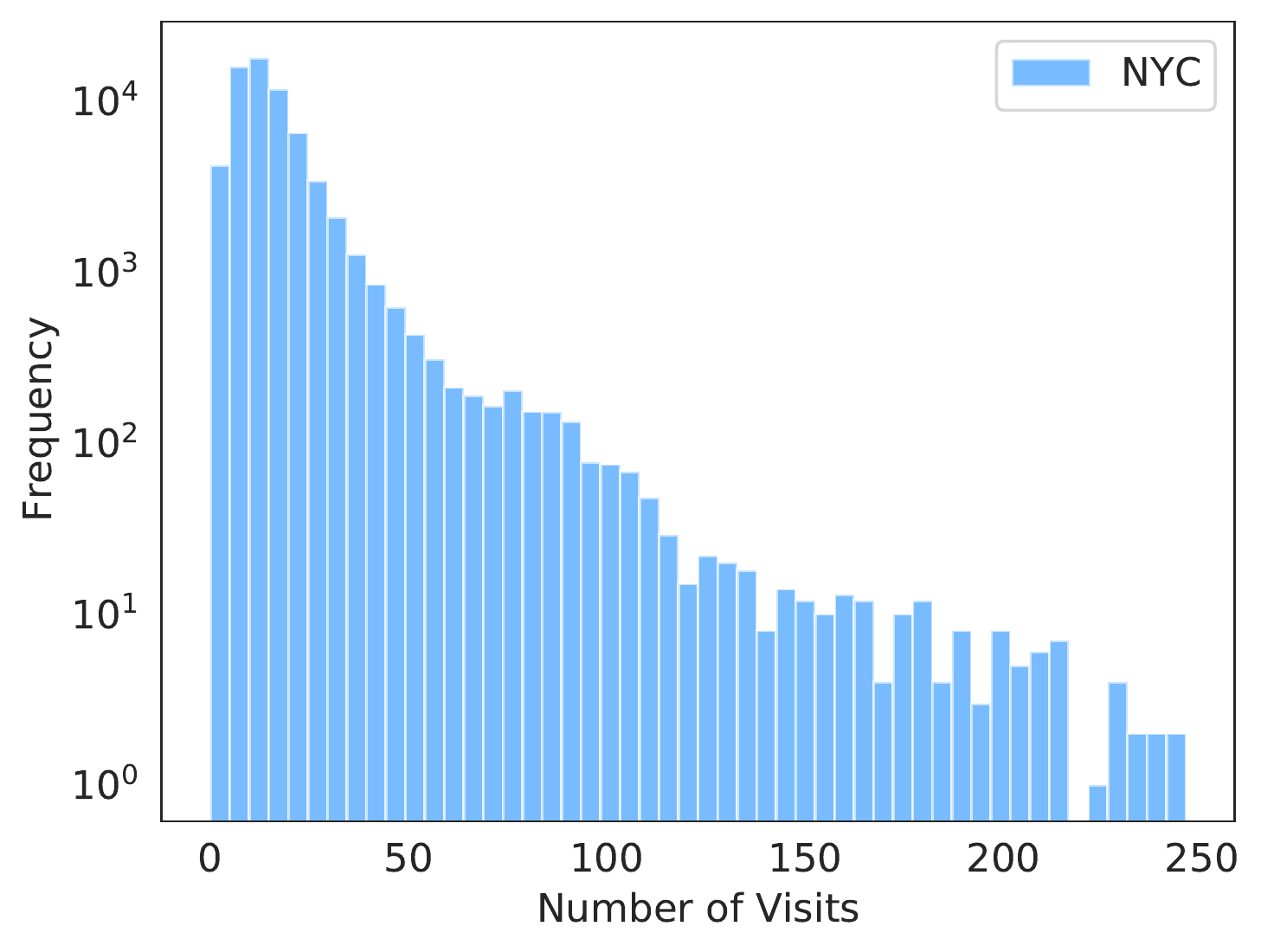}} 
    \subfigure[Dallas]{
    \includegraphics[width=.3\textwidth]{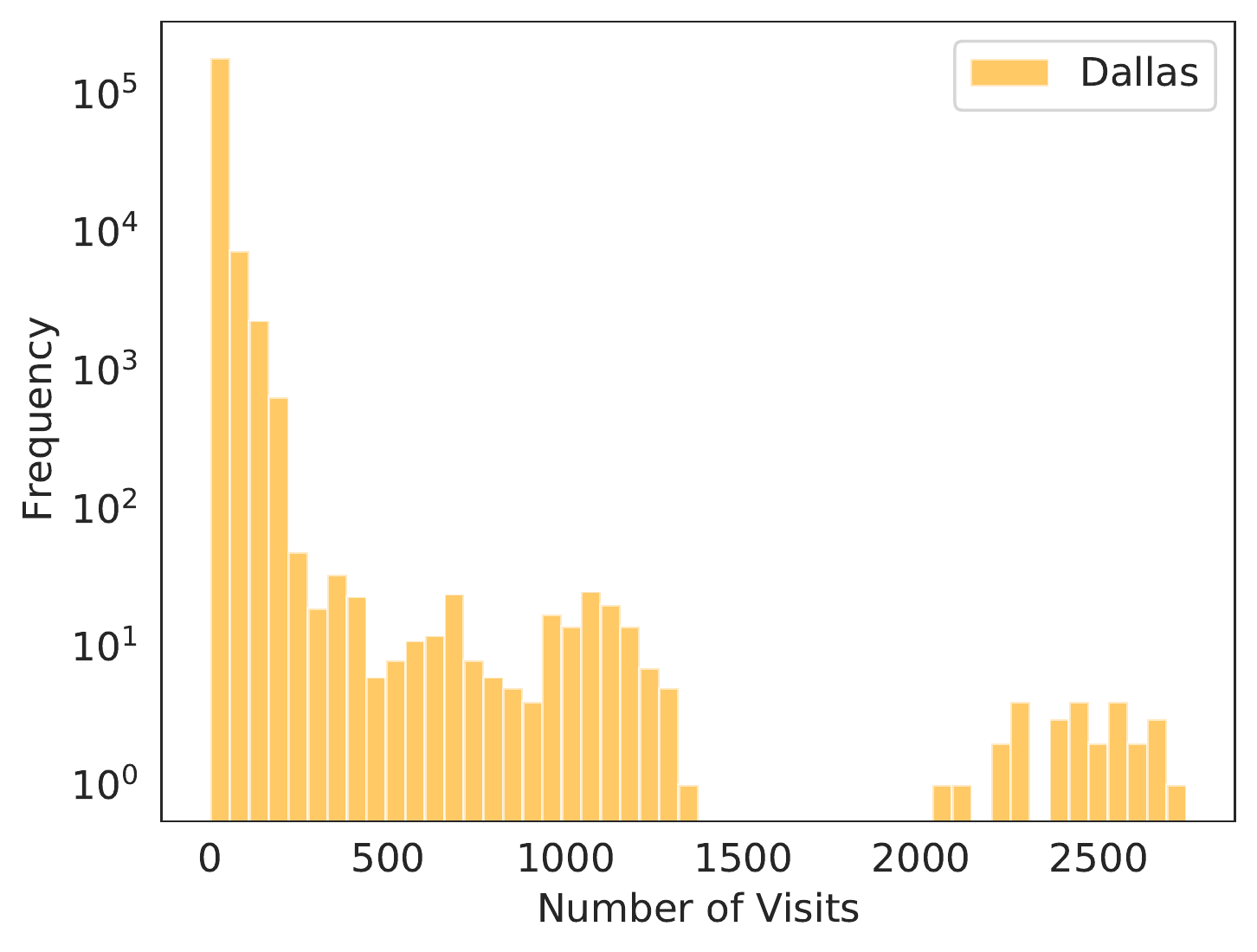}} 
    \subfigure[Miami]{
    \includegraphics[width=.3\textwidth]{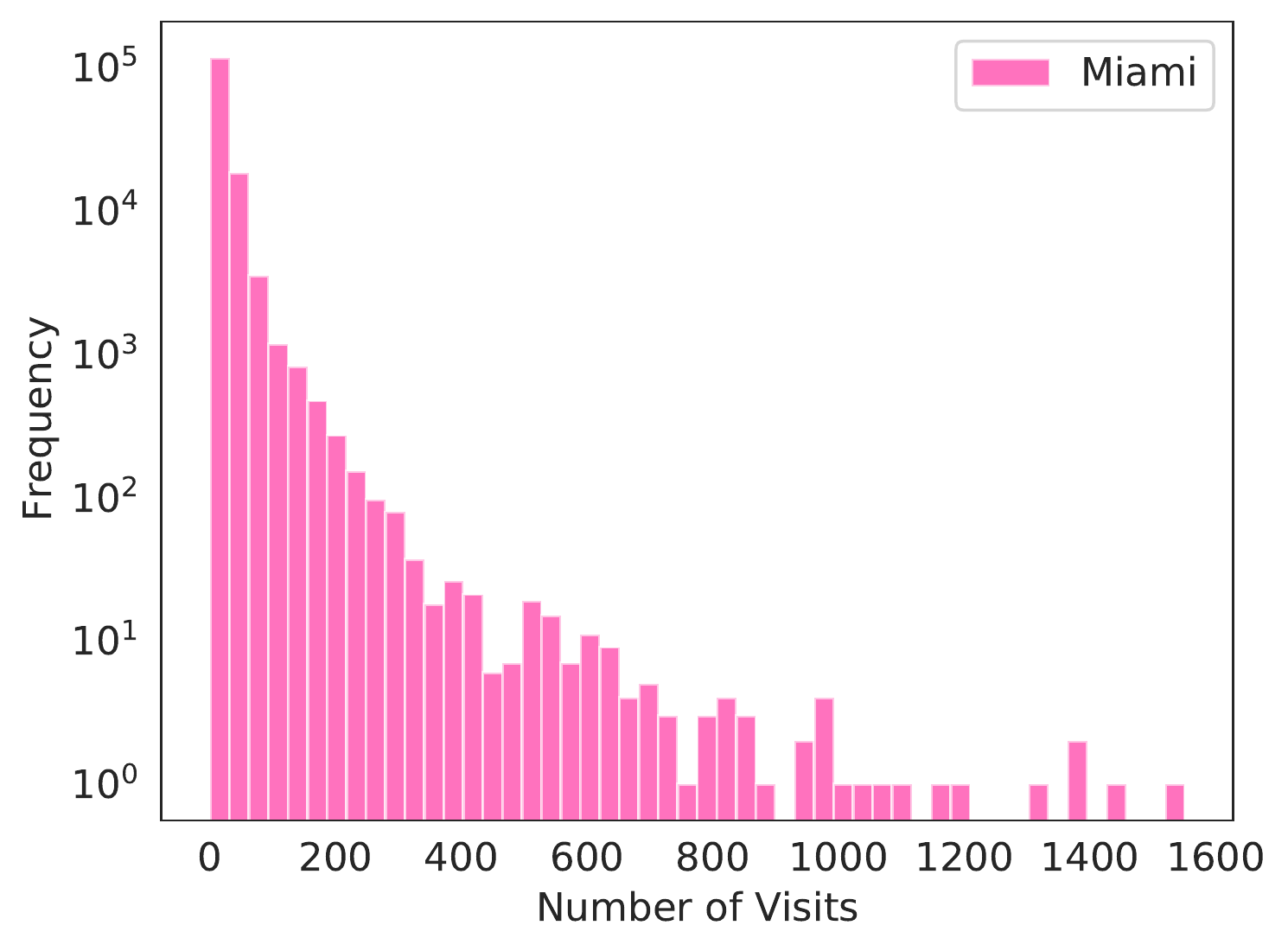}} 
    \caption{The distribution of the number of visits in three datasets.}
    \label{fig:dis}
\end{figure}

In addition to the statistics of three datasets listed in Table~\ref{tab:dataset}, the distribution plots of the number of visits are presented in Figure~\ref{fig:dis}.
Based on these plots, we can see that the distribution of three cities are different.
As shown in the main paper, the proposed \name\ achieves good forecasting results on all three datasets, which further demonstrates the robustness of our \name.

\subsection{Computational Cost}\label{sec:cost}

\begin{table}[]
\centering
\caption{ Comparison of computational cost.}
\begin{tabular}{l|cc} \hline
 & GPU memory (MB) & \# Parameters \\ \hline
Gru & 879 & 0.396$\times 10^6$ \\
GruA & 911 & 0.922$\times 10^6$ \\
Transformer & 1009 & 3.194$\times 10^6$ \\
Reformer & 1077 & 4.342$\times 10^6$ \\
Informer & 1007 & 4.410$\times 10^6$ \\ \hline
S2S(GruA) & 1450 & 3.919$\times 10^6$ \\
S2S(Transformer) & 1455 & 4.898$\times 10^6$ \\
S2S(BART) & 6743 & 139.42$\times 10^6$ \\ \hline
SHIFT (Ours) & 1469 & 4.898$\times 10^6$ \\ \hline
\end{tabular}
\label{tab:cost}
\end{table}
In this section, we analysis the computational cost of \name.
Table~\ref{tab:cost} lists the GPU memory usage (in MB) during training and the number of trainable parameters of each method.
These statistics are benchmarked while training each model on the NYC dataset.
Generally, the computational cost of language-based models (both \pname\ and \name) are larger than numerical value-based forecasting models.
Also, Transformer-based models require more resources than GRU-based models.
These two observations are as expected.
Among all language-based methods, the cost of our \name\ is very close and comparable to \pname(GruA) and \pname(Transformer), whereas the cost of \pname(BART) is significantly larger than others.
From the table, we also notice that the number of trainable parameters of \name\ is almost the same as \pname(Transformer) while \name\ takes a little bit more GPU memory.
Due to the extra \mb\ branch in \name, it takes more memory during training.
However, since the \mb\ branch encoder is updated in the momentum mode, this branch does not introduce many trainable parameters.

\subsection{Pseudo-code of \name}
In Algorithm~\ref{alg}, the pseudo-code of \name\ training process (using one epoch as example) is presented in the PyTorch-like style.

\begin{algorithm}[]
\caption{Pseudo-code of training \name\ (PyTorch-like)}
\label{alg}
\textcolor{gray}{\# $\phi_n$, $f_N$, $f_D$: the embedding layer, encoder, and decoder of the \nl\ branch}
\\
\textcolor{gray}{\# $\phi_m$, $f_M$, $\text{MLP}$: the embedding layer, encoder, and predictor of the \mb\ branch}
\\
\textcolor{gray}{\# $\alpha_{loss}$, $\alpha_m$: loss factor and momentum factor}
\begin{algorithmic}[1]
\State{$\theta_M = \theta_N$} \textcolor{gray}{\Comment{Momentum updating initialization}}
\For{$(\mathbf{X}, Y, {x}_{t_{\text{obs} + 1}}, \hat{Y})$ in train\_data\_loader} \textcolor{gray}{\Comment{Loading a batch of training data}}
\State{$h_N = f_N (\phi_n (Y))$} \textcolor{gray}{\Comment{\nl\ branch encoding, Eqs.~\eqref{eq:nl_emb} \&~\eqref{eq:nl_enc}}}
\State{$\hat{Y} = f_D (h_N)$} \textcolor{gray}{\Comment{\nl\ branch decoding, Eq.~\eqref{eq:nl_pred}}}
\State{$h_M = f_M (\phi_m (\mathbf{X}))$} \textcolor{gray}{\Comment{\mb\ branch encoding, Eqs.~\eqref{eq:mb_emb} \&~\eqref{eq:mb_enc}}}
\State{$\tilde{x}_{t_{\text{obs} + 1}} = \text{MLP}(h_M)$} \textcolor{gray}{\Comment{\mb\ branch prediction, Eq.~\eqref{eq:mb_out}}}
\State{$\mathcal{L}_N = \text{CrossEntropy}(\hat{Y}, Y)$} \textcolor{gray}{\Comment{\nl\ branch loss, Eq.~\eqref{eq:nl_loss}}}
\State{$\mathcal{L}_{M} = \text{MSE}(\tilde{x}_{t_{\text{obs} + 1}}, {x}_{t_{\text{obs} + 1}})$} \textcolor{gray}{\Comment{\mb\ branch loss, Eq.~\eqref{eq:mb_loss}}}
\State{$\mathcal{L}= (1 - \alpha_{loss}) \mathcal{L}_{N} + \alpha_{loss} \mathcal{L}_{M}$} \textcolor{gray}{\Comment{Total loss, Eq.~\eqref{eq:loss}}}
\State{$\mathcal{L}.\text{backward}()$} \textcolor{gray}{\Comment{Back propagation}}
\State{$\text{update}(\text{\name}.\text{params})$} \textcolor{gray}{\Comment{Update \name\ parameters except for $\theta_M$}}
\State{$\theta_{M} \leftarrow \alpha_{m}\theta_{N} + (1 - \alpha_{m}) \theta_{M}$} \textcolor{gray}{\Comment{Momentum updating \mb\ branch, Eq.~\eqref{eq:momentum}}}
\EndFor
\end{algorithmic}
\end{algorithm}

\end{document}